\documentclass[11pt,a4paper]{article}
\usepackage[nohyperref]{naaclhlt2019}
\usepackage{times}
\usepackage{latexsym}
\usepackage{soul}

\usepackage{url}
\usepackage{boxedminipage}
\usepackage{xcolor}
\usepackage{amsmath}
\usepackage{bm}
\usepackage{xspace}
\usepackage{enumitem}
\usepackage{cuted}

\aclfinalcopy 

\usepackage{multirow}
\usepackage{booktabs}

\usepackage{graphicx} 
\graphicspath{ {figures/} }
\usepackage{subcaption}
\usepackage{rotating}

\usepackage{txfonts}

\newcommand{\jfleg}{\textsc{JFLEG}\xspace}
\newcommand{\jfleges}{\textsc{JFLEG-es}\xspace}
\newcommand{\nucle}{\textsc{NUCLE}\xspace}
\newcommand{\wmt}{\textsc{WMT}\xspace}
\newcommand{\gec}{\textsc{GEC}\xspace}
\newcommand{\clean}{\textsc{clean}\xspace}
\newcommand{\drop}{\textsc{drop}\xspace}
\newcommand{\art}{\textsc{art}\xspace}
\newcommand{\prep}{\textsc{prep}\xspace}
\newcommand{\nn}{\textsc{nn}\xspace}
\newcommand{\sva}{\textsc{sva}\xspace}
\newcommand{\dropd}{\textsc{clean+drop}\xspace}
\newcommand{\artd}{\textsc{clean+art}\xspace}
\newcommand{\prepd}{\textsc{clean+prep}\xspace}
\newcommand{\nnd}{\textsc{clean+nn}\xspace}
\newcommand{\svad}{\textsc{clean+sva}\xspace}
\newcommand{\mixd}{\textsc{mix-both}\xspace}
\newcommand{\mixall}{\textsc{mix-all}\xspace}

\newcommand{\rf}[1]{\textsc{cor#1}\xspace}

\newcommand{\emp}{\ensuremath{\emptyset}}

\newcommand\blfootnote[1]{%
  \begingroup
  \renewcommand\thefootnote{}\footnote{#1}%
  \addtocounter{footnote}{-1}%
  \endgroup
}

\title{Neural Machine Translation of Text from Non-Native Speakers}

\author{Antonios Anastasopoulos$^\dagger$ \qquad Alison Lui$^\dagger$ \qquad Toan Nguyen \qquad David Chiang \\
  Department of Computer Science and Engineeering\\
  University of Notre Dame \\
 {\tt \{aanastas,alui,tnguye28,dchiang\}@nd.edu}
 }
\date{}

\begin{document}
\maketitle
\blfootnote{$^\dagger$Equal contribution.}
\begin{abstract}
Neural Machine Translation (NMT) systems are known to degrade when confronted with noisy data, especially when the system is trained only on clean data. In this paper, we show that augmenting training data with sentences containing artificially-introduced grammatical errors can make the system more robust to such errors. In combination with an automatic grammar error correction system, we can recover $1.9$ BLEU out of $3.1$ BLEU lost due to grammatical errors.
We also present a set of Spanish translations of the JFLEG grammar error correction corpus, which allows for testing NMT robustness to real grammatical errors.
\end{abstract}

\section{Introduction}

Neural Machine Translation (NMT) is undeniably a success story: public benchmarks \cite{bojar2016findings} are dominated by neural systems, and neural approaches are the \emph{de facto} option for industrial systems \cite{wu2016google,humanparity,crego2016systran,hieber2018sockeye}. Even under low-resource conditions, neural models were recently shown to outperform traditional statistical approaches \cite{nguyen+chiang:naacl2018}.

However, there are still several shortcomings of NMT that need to be addressed: a (non-exhaustive) list of six challenges is discussed by \citet{koehn2017six}, including out-of-domain testing, rare word handling, the wide-beam problem, and the large amount of data needed for learning. An additional challenge is robustness to noise, both during training and at inference time. 

In this paper, we study the effect of a specific type of noise in NMT: grammatical errors.
We primarily focus on errors that are made by non-native source-language speakers (as opposed to dialectal language, SMS or Twitter language). Not only is this linguistically important, but we believe that it would potentially have great social impact.

Our contributions are three-fold.
First, we confirm that NMT is vulnerable to source-side noise when trained on clean data, losing up to $3.6$ BLEU on our test set. This is consistent with previous work, yet orthogonal to it, since we use more realistic noise for our experiments.
Second, we explore training methods that can deal with noise, and show that including noisy synthetic data in the training data makes NMT more robust to handling similar types of errors in test data.
Combining this simple method with an automatic grammar correction system, we find that we can recover $1.5$ BLEU.
Third, we release Spanish translations of the JFLEG corpus,\footnote{Freely available at \ifaclfinal\url{https://bitbucket.com/antonis/nmt-grammar-noise}\else \url{URL} \fi} a standard benchmark for English Grammar Error Correction (GEC) systems. 
We also release all other data and code used in this paper.

Our additional annotations on both the \jfleg corpus and the English \wmt data will enable the evaluation of the robustness of NMT systems on realistic, natural noise:
a robust system would ideally produce the same output when presented with either the original or the noisy source sentence.
We hope that our datasets will become a benchmark for noise-robust NMT, because we believe that deployed systems should also be able to handle source-side noise.

\section{Data}

We focus on NMT from English to Spanish. We choose English to be our source-side language because there exist English corpora annotated with grammar corrections, which we can use as a source of natural noise. Moreover, since English is probably the most commonly spoken non-native language \cite{lewis2009ethnologue}, our work could be directly applicable to several translation applications. Our choice of Spanish as a target language enables us to have access to existing parallel data and easily create new parallel corpora (see below, \S\ref{sec:jfleg}).  

For all experiments, we use the Europarl English-Spanish dataset \cite{koehn2005europarl} as our training set. In the synthetic experiments of Section~\S\ref{sec:synth}, we use the newstest2012 and newstest2013 as dev and test sets, respectively. Furthermore, to test our translation methods on real grammatical errors, we introduce a new collection of Spanish translations of the JFLEG corpus~(\S\ref{sec:jfleg}).

\subsection{Grammar Error Correction Corpora}
\label{sec:gec}

To our knowledge, there are five publicly available corpora of non-native English that are annotated with corrections, which have been widely used for research in Grammar Error Correction (\textsc{GEC}). 
The NUS Corpus of Learner English (\textsc{NUCLE}) contains essays written by students at the National University of Singapore, corrected by two annotators using 27 error codes \cite{dahlmeier2013building}. It has become the main benchmark for GEC, as it was used in the CoNLL GEC Shared Tasks \cite{ng2013conll,ng2014conll}. Other corpora include the Cambridge Learner Corpus First Certificate in English \textsc{FCE} corpus \cite{yannakoudakis2011new}, which is only partially public, the \textsc{Lang-8} corpus \cite{tajiri2012tense}, which was harvested from online corrections, and the \textsc{AESW} 2016 Shared Task corpus, which contains corrections on texts from scientific journals.

The last corpus
is the JHU FLuency-Extended GUG corpus (\textsc{JFLEG}) \cite{napoles-sakaguchi-tetreault:2017:EACLshort}. 
This corpus covers a wider range of English proficiency levels on the source side, and its correction annotations include extended fluency edits rather than just minimal grammatical ones.
That way, the corrected sentence is not just grammatical, but also guaranteed to be fluent.

\subsection{Synthetic grammar errors}
\label{sec:synth}
\begin{table}[t]
    \centering
    \begin{tabular}{c|p{4cm}}
    \toprule
        Error Type & Confusion Set \\
    \midrule
        \art & \{a, an, the, \emp \} \\
        \prep & \{on, in, at, from, for, under, over, with, into, during, until, against, among, throughout,of, to, by, about, like, before, after, since, across, behind, but, out, up, down, off, \emp\}\\
        \nn & \{SG, PL\}\\
        \sva & \{3SG, not 3SG, 2SG-Past, not 2SG-Past\}\\
    \bottomrule
    \end{tabular}
    \caption{Confusion sets for each grammar error type. The \art and \prep sets include an empty token (\emp) allowing for insertions and deletions. SG, PL, 2SG, and 3SG stand for singular, plural, second-person and third-person singular respectively.}
    \label{tab:confusion}
\end{table}
\begin{table}[t]
    \centering
    \begin{tabular}{cccc}
    \toprule
        \multirow{2}{*}{Dataset} & \multicolumn{3}{c}{Percentage of Errors} \\
         & Train & Dev & Test \\
    \midrule
        sentences & 2M & 3K & 3K \\
        words & 55M & 74K & 73K \\
    \midrule
        \drop & 100\% & 100\% & 100\% \\
        \art & 96.4\% & 98.4\% & 99.8\% \\
        \prep & 95.7\% & 95.9\% & 98.4\% \\
        \nn & 94.5\% & 91.0\% & 98.6\% \\
        \sva & 93.1\% & 81.9\% & 82.0\% \\
    \midrule
        \dropd & 50\% & 50\% & -- \\
        \artd & 48.2\% & 49.1\% & -- \\
        \prepd & 47.8\% & 47.9\% & -- \\
        \nnd & 47.3\% & 45.5\% & -- \\
        \svad & 46.5\% & 41.0\% & -- \\
    \midrule
        \mixall & 39.9\% & 38.9\% & -- \\
    \bottomrule
    \end{tabular}
    \caption{Statistics on the original and synthetic En-Es datasets. Each (synthetic) sentence has exactly one introduced error, wherever possible.  
    \textsc{clean+[error]} is the concatenation of the \textsc{[error]} with the original clean dataset, while
    \mixall includes six versions of each training sentence, one without errors and one for each error.}
    \label{tab:synthdata}
\end{table}
\begin{table*}[t]
    \centering
    \begin{tabular}{c|p{13cm}}
    \toprule
        Error Type & Example \\
    \midrule
        \multirow{4}{*}{\art} & In October , Tymoshenko was sentenced to seven years in prison for entering into what was reported to be \emph{a/*}{\emp} disadvantageous gas deal with Russia.\\
         &  Its ratification would require \emp\emph{/*the} 226 votes.\\
         & It is \emph{a/*the} good result, which nevertheless involves a certain risk.\\
    \midrule 
        \multirow{3}{*}{\prep} & [\ldots] the motion to revoke an article based \emph{on/*in} which the opposition leader , Yulia Tymoshenko , was sentenced.\\
        &  Its ratification would require \emp\emph{/*for} 226 votes.\\
    \midrule
        \multirow{3}{*}{\nn} & Its ratification would require 226 \emph{votes/*vote}. \\
         & The \emph{verdict/*verdicts} is not yet final ; the court will hear Tymoshenko 's appeal in December. \\
    \midrule
        \multirow{3}{*}{\sva} & As a rule, Islamists \emph{win/*wins} in the country; the question is whether they are the moderate or the radical ones. \\
        & This cultural signature \emph{accompanies/*accompany} the development of Moleskine;\\
    \bottomrule
    \end{tabular}
    \caption{Example grammatical errors that were introduced in the En-Es WMT test set.}
    \label{tab:exampleerrors}
\end{table*}

Ideally, we would train a translation model to translate grammatically noisy language by training it on parallel data with grammatically noisy language. Since, to our knowledge, no such data exist in the quantities that would be needed, an alternative is to add synthetic grammatical noise to clean data. 
An advantage of this approach is that controlled introduction of errors allows for fine-grained analysis.

This is a two-step process, similar to the methods used in the \gec literature for creating synthetic data based on confusion matrices \cite{rozovskaya2014illinois,rozovskaya2010generating,xie2016neural,sperbertoward}.
First, we mimic the distribution of errors found in real data, and then we introduce errors by applying rule-based transformations on automatic parse trees.

The first step involves collecting error statistics on real data.
Conveniently, the \nucle corpus has all corrections annotated with 27 error codes. 
We focus on five types of errors, with the last four being the most common in the NUCLE corpus:
\begin{itemize}[nolistsep]
    \item \drop: randomly deleting one character from the sentence.\footnote{This error is not part of the NUCLE error list.}
    \item \art: article/determiner errors
    \item \prep: preposition errors
    \item \nn: noun number errors
    \item \sva: subject-verb agreement errors
\end{itemize}

Using the annotated training set of the \nucle corpus, we compute error distribution statistics, resulting in confusion matrices for the cases outlined in Table~\ref{tab:confusion}. 
For \art and \prep errors, we obtain probability distributions that an article, determiner, or preposition is deleted, substituted with another member of the confusion set, or inserted in the beginning of a noun phrase. 
For \nn errors, we obtain the probability of a noun being replaced with its singular or plural form. For \sva errors, the probability that a present tense verb is replaced with its third-person-singular (3SG) or not-3SG form. An additional \sva error that we included is the confusion between the appropriate form for the verb `to be' in the past tense (`was' and `were').

The second step involves applying the noise-inducing transformations using our collected statistics as a prior.
We obtained parses for each sentence using the Berkeley parser \cite{petrov2006learning}.
The parse tree allows us to identify candidate error positions in each sentence (for example, the beginning of a noun phrase without a determiner, were one could be inserted).
For each error type we introduced exactly \emph{one} error per sentence, wherever possible, which we believe matches more realistic scenarios than previous work.
It also allows for controlled analysis of the behaviour of the NMT system (see Section~\ref{sec:analysis}).

For each error and each sentence, we first identify candidate positions (based on the error type and the parse tree) and sample one of them based on the specific error distribution statistics.
Then, we sample and introduce a specific error using the corresponding probability distribution from the confusion matrix.
(In the case of \textsc{drop}, \textsc{nn}, and \textsc{sva} errors, we only need to sample the position and only insert/substitute the corresponding error.)
If no candidate positions are found (for example, a sentence doesn't have a verb that can be substituted to produce a \sva error) then the sentence remains unchanged.

Following the above procedure, we added errors in our training, dev, and test set (henceforth referred to as \textsc{[error]}).
Basic statistics on our produced datasets can be found in Table~\ref{tab:synthdata}, while example sentences are shown in Table~\ref{tab:exampleerrors}. 
Furthermore, we created training and dev sets that mix clean and noisy data. The \textsc{clean+[error]} training sets are the concatenation of each \textsc{[error]} with the clean data, effectively including a clean and a noisy version of each sentence pair.

We also created a training and dev dataset with mixed error types, in our attempt to study the effect of including all noise types during training. 
The \mixall dataset includes each training pair six times: once with the original (clean) sentence as the source, and once for every possible error.
We experimented with a mixed dataset that included each training sentence once, with the number of noisy sentences being proportional to the real error distributions of the \nucle dataset,
but obtained results similar to the \textsc{[error]} datasets.

\subsection{\jfleges: Spanish translations of JFLEG}
\label{sec:jfleg}

The \jfleg corpus consists of a dev and test set (no training set), with 747 and 754 English sentences, respectively, collected from non-native English speakers. 
Each sentence is annotated with four different corrections, resulting in four (fluent and grammatical) reference sentences. About
$14\%$ of the sentences do not include any type of error, with the source and references being equivalent.

We created translations of the \jfleg corpus that allow us to evaluate how well NMT fares compared to a human translator, when presented with noisy input.
We will refer to the augmented \jfleg corpus as \jfleges.

Two professional translators were tasked with producing translations for the dev and the test set, respectively. The translators were presented only with the original erroneous sentences; they did not have access to the correction annotations. 
They were asked to produce fluent, grammatical translations in European Spanish (to match the Spanish used in the Europarl corpus).
There exist cases where a translator might choose to preserve a source-side error when producing the translation, such as translation of literary works where it's possible that grammar or fluency errors are intentional; however, our translators were explicitly asked not to do that. The exact instructions were as follows:
\begin{quote}
    Please translate the following sentences. Note that some sentences will have grammatical errors or typos in English. Don't try to translate the sentences word for word (e.g. replicate the error in Spanish). Instead, try to translate it as if it was a grammatical sentence, and produce a fluent grammatical Spanish sentence that captures its meaning.
\end{quote}

\section{Experiments}

In this section, we provide implementation details and the results of our NMT experiments. For convenience, we will refer to each model with the same name as the dataset it was trained on; e.g. the \mixall model will refer to the model trained on the \mixall dataset.

\subsection{Implementation Details}

All data are tokenized, truecased, and split into subwords using Byte Pair Encoding (BPE) with $32{,}000$ operations \cite{sennrich2016}. 
We filter the training set to contain only sentences of up to 80 words.

Our LSTM models are implemented using DyNet \cite{neubig2017dynet}, and our transformer models using PyTorch \cite{paszke2017automatic}.
The transformer model uses 4 layers, 4 attention heads, the dimension for embeddings and positional feedforward are 512 and 2048 respectively . The sublayer computation sequence follows the guidelines from \citet{chen2018best}.
Dropout is set to 0.8 (also in the source embeddings, following \citet{sperbertoward}). We use the learning rate schedule in \citet{transformer} with warm-up steps of 24000 but only decay the learning rate until it reaches $e^{-4}$ as inspired by \citet{chen2018best}.
For testing, we select the model with the best performance on the dev set corresponding to the test set.
At inference time, we use a beam size of 4 with length normalization \cite{wu2016google} with a weight of $0.6$.  


\subsection{Results}
\label{sec:results}

\begin{table*}[t]
\newcommand{\match}[1]{\hl{#1}}
    \centering
    \begin{tabular}{c|c|ccccc|c}
    \toprule
        \multirow{2}{*}{WMT Training Set} &  \multicolumn{7}{c}{En-Es WMT Test Set}\\ 
         &  \clean & \drop & \art & \prep & \nn &\sva & \textsc{average} $\pm$ \textsc{stdev} \\
        \midrule
        \clean & \textbf{\match{33.0}} & 29.6 & 31.3 & \textbf{32.0} & 29.3 & \textbf{32.1} & $31.2 \pm 1.5$\\
    \midrule
        \drop & 31 & \match{30.2} & 30.0 & 30.0 & 28.3 & 30.6 & $30.0 \pm 0.9$\\
        \art  & 31.2 & 28.4 & \match{30.8} & 30.2 & 27.7 & 30.8 & $29.8 \pm 1.4$\\
        \prep & 30.4 & 27.8 & 29.3 & \match{30.3} & 27.4 & 29.9 & $29.2 \pm 1.3$\\
        \nn   & 30.4 & 27.9 & 28.9 & 29.5 & \match{29.8} & 29.8 & $29.4 \pm 0.8$\\ 
        \sva  & 31.2 & 28.7 & 30.2 & 30.3 & 28.2 & \match{30.9} & $29.9 \pm 1.2$\\ 
    \midrule
        \dropd & \textbf{32.9} & \textbf{\match{31.4}} & \textbf{31.4} & 31.8 & 29.5 & \textbf{32.0} & $31.5 \pm 1.2$\\
        \artd  & \textbf{32.7} & 29.7 & \textbf{\match{31.7}} & 31.7 & 28.8 & \textbf{32.1} & $31.1 \pm 1.5$\\
        \prepd & \textbf{32.7} & 29.6 & 31.2 & \textbf{\match{32.2}} & 29.0 & 31.8 & $31.1 \pm 1.5$\\
        \nnd & 32.5 & 29.4 & 30.7 & 31.4 & \textbf{\match{31.0}} & 31.6 & $31.1 \pm 1.0$\\ 
        \svad  & 32.5 & 29.6 & 31.2 & 31.5 & 29.0 & \textbf{\match{31.9}} & $30.9 \pm 1.4$\\ 
    \midrule
        \mixall & \textbf{32.7} & 30.9 & \textbf{31.4} & \textbf{32.0} & \textbf{30.6} & \textbf{32.0} & \textbf{31.6} $\pm$ \textbf{0.7} \\
    \bottomrule
    \end{tabular}
    \caption{BLEU scores on the WMT test set without (\textsc{clean}) and with synthetic grammar errors. The best performing models for each test set are \textbf{highlighted}. When training and test match (\match{highlighted}) we generally observe higher results.
    However, including all clean and noisy data in the training set (\textsc{mix-all}) yields the best results across all datasets, with the highest average BLEU \emph{and} the lowest variance.}
    \label{tab:results}
\end{table*}

We report the results obtained with the transformer model, as they were consistently better than the LSTM one. All the result tables and corresponding analysis for the LSTM models can be found in the Supplementary Material.

The performance of our systems on the synthetic WMT test sets, as measured by detokenized BLEU \cite{papineni2002bleu}, is summarized in Table~\ref{tab:results}. When the system is trained only on clean data (first row) and tested on noisy data, it unsurprisingly exhibits degraded performance. We observe significant drops in the range of $1.0$--$3.6$ BLEU. 

The largest drop (more than~3 BLEU) is observed with \nn errors in the source sentence. This is not unreasonable: nouns almost always carry content significant for translation. Especially when translating into Spanish, a noun number change can, and apparently does, also affect the rest of the sentence significantly, for example, by influencing the conjugation of a subsequent verb.
The second-largest drop (more than~$2.5$ BLEU points) is observed in the case of \drop errors. This is also to be expected; typos produce out-of-vocabulary (OOV) words, which in the case of BPE are usually segmented to a most likely rarer subword sequence than the original correct word.

We find that a training regime that includes both clean and noisy sentences (\textsc{[clean+error}) results in better systems across the board. Importantly, these models manage to perform en par with the \clean model on the \clean test set.
Since the original training set is part of the \textsc{[clean+error} training sets, this behavior is expected. We conclude, thus, that including the full clean dataset during training is important for performance on clean data -- one cannot just train on noisy data.

The \textsc{[clean+error]} systems exhibit a notable pattern: their BLEU scores are generally similar to the \clean system on all test sets, except for the test set that matches their training set errors (highlighted in Table~\ref{tab:results}), where they generally obtain the best performance. 

The \mixall model is our best system on all test sets and on average. Unlike the \textsc{[clean+error]} systems, it outperforms the \clean model on \emph{all} noisy test sets and not only on a specific one.
On average, using the \mixall training set leads to an improvement of $0.7$ BLEU over the \clean model and $0.3-0.8$ BLEU over the \textsc{[clean+error]} models.
Furthermore, the \mixall model is the one with the smallest performance standard deviation out of all the models, when averaging over all the test sets. 
This is another indication that our system is more robust to multiple source-side variations. We further explore this intuition in Section~\ref{sec:analysis}.

\begin{table*}[t]
    \centering
    \jfleges Dev \\[1ex]
    \begin{tabular}{c|ccccc|cc}
    \toprule
        \multirow{2}{*}{Training} &  \multicolumn{5}{c|}{Manual correction} & No & Auto \\ 
         & \rf0 & \rf1 & \rf2 & \rf3 & avg. & corr. & corr. \\
    \midrule
        \clean & 32.1 & 31.5 & 32.5 & 33.3 & 32.4 & 31.1 & 31.2 \\
        \mixall  & 31.9 & 31.4 & 32.2 & \textbf{32.9} & \textbf{32.1} & \textbf{32.2} & \textbf{31.6}\\
    \bottomrule
    \end{tabular}
    
    \vspace*{2ex}
    \jfleges Test \\[1ex]
    \begin{tabular}{c|ccccc|cc}
    \toprule
        \multirow{2}{*}{Training} &  \multicolumn{5}{c|}{Manual correction} & No & Auto \\ 
         & \rf0 & \rf1 & \rf2 & \rf3 & avg. & corr. & corr. \\
    \midrule
    \clean & \textbf{28.4} & \textbf{28.8} & \textbf{29.1} & \textbf{28.2} & \textbf{28.6} & 26.2 & \textbf{27.0} \\
    \mixall & 27.7 & 28.1 & 28.1 & 27.5 & 27.8  & \textbf{26.8} & \textbf{26.7} \\
    \bottomrule
    \end{tabular}
    \caption{BLEU scores on the \jfleges dev and test datasets. Our proposed \mixall model is comparable to the \clean model on manually corrected input (\textsc{cor[0--3]}). On noisy input (No corr.) the \mixall outperforms the \clean model ($25.6 > 24.2$).
    Preprocessing the noisy input with a GEC model (Auto corr.) further improves results ($26.1$).}
    \label{tab:jfleg-results-test}
\end{table*}

\fi

On the more realistic \jfleges dev and test sets, we observe same trends but at a smaller scale, as shown in Table~\ref{tab:jfleg-results-test}. Our \mixall model generally achieves comparable results when presented with each of the four reference corrections of the test set (\textsc{corX} columns).
However, when we use the noisy source sentence as input (\textsc{No corr} column) our \mixall model obtains $1.4$ BLEU improvements over the \clean model.
The \emph{difference} between the performance of the models when presented with clean and noisy input is another indicator for robustness. On the \jfleges test set, the noisy source results in a $-3.1$ BLEU point drop for the \clean model, while the drop for our \mixall model is smaller, at $-1.7$ BLEU points.

In addition, we experimented with using an automatic error-corrected source as input to our system (column \textsc{Auto corr} of Table~\ref{tab:jfleg-results-test}). 
We used the publicly available \jfleg outputs of the (almost) state-of-the-art model of   \citet{junczysdowmunt-grundkiewicz:2016:EMNLP2016} as inputs to our NMT system.\footnote{This model has been recently surpassed by other systems, e.g. \cite{dowmunt2018etal}, but their outputs are not available online.}
This experiment envisions a pipeline where the noisy source is first automatically corrected and then translated. As expected, this helps the \clean model (by $+1.1$ BLEU), but our \mixall training helps even further (by another $+0.8$ BLEU). Interestingly, the automatic GEC system only helps in the test set, while there are no improvements in the dev set.
Naturally, since automatic GEC systems are imperfect, the performance of this pipeline still lags behind translating on clean data.

\section{Analysis}
\label{sec:analysis}

We attempt an in-depth analysis of the impact of the different source-side error types on the behavior of our NMT system, when trained on clean data and tested on the artificial noisy data that we created.

\paragraph{\textsc{Art} Errors} Table~\ref{tab:art} shows the difference of the BLEU scores obtained on the sentences, broken down by the type of article error that was introduced. The first observation is that in all cases the difference is negative, meaning that we get higher BLEU scores when testing on clean data. 
Encouragingly, there is practically no difference when we substitute `a' with `an' or `an' with `a'; the model seems to have learned very similar representations for the two indefinite articles, and as a result such an error has no impact on the produced output.
However, we observe larger performance drops when substituting indefinite articles with the definite one and vice versa; 
since the target language makes the same article distinction as the source language, any article source error is propagated to the produced translation.

\paragraph{\textsc{Prep} Errors} Due to the large number of prepositions, we cannot present a full analysis of preposition errors, but highlights are shown in Table~\ref{tab:prep}. Deleting a correct preposition or inserting a wrong one leads to performance drops of $1.2$ and $0.8$ BLEU points for the \clean model, but drops of $0.4$ and $0.7$ for the \mixall model.

\paragraph{\textsc{Nn} and \textsc{Sva} Errors} We found no significant performance difference between the different \nn errors. Incorrectly pluralizing a noun has the same adverse effect as singularizing it, leading to performance reductions of over $4.0$ and $3.5$ BLEU points respectively. We observe a similar behavior with \sva errors: each error type leads to roughly the same performance degradation.

\begin{table}[t]
    \centering
    \begin{tabular}{c|ccccc}
    \toprule
        Correct & \multicolumn{5}{c}{Substituted article}  \\
         article & a & an & the & {\emp} & \multicolumn{1}{c}{\emph{all}} \\
    \midrule
        a & -- & 0 & $-2.0$ & $-2.1$ & $-2.1$ \\
        an & 0 & -- & $-5.7$ & $-7.3$ & $-6.3$  \\
        the & $-4.1$ & $-2.2$ & -- & $-1.7$ & $-1.8$ \\
        {\emp} & $-3.1$ & $-3.7$ & $-1.5$  & -- & $-1.7$ \\
        \emph{all} & $-3.8$ & $-3.4$ & $-1.5$ & $-1.8$ & $-1.7$ \\
    \bottomrule
    \end{tabular}
    \caption{Effect of article substitutions in test data (\art{}) relative to clean test data (\clean{}), broken down by substitution type. Different article substitutions have very different impacts on BLEU; changing an indefinite article to definite is especially damaging.}
    \label{tab:art}
\end{table}

\section{Related Work}

The effect of noise in NMT was recently studied by \citet{khayrallah-koehn:2018:WNMT}, who explored noisy situations during training due to web-crawled data. 
This type of noise includes misaligned, mistranslated, or untranslated sentences which, when used during training, significantly degrades the performance of NMT. 
Unlike our work, they primarily focus on a setting where the training set is noisy but the test set is clean.

In addition, \citet{heigold2017robust} evaluated the robustness of word embeddings against word scrambling noise, and showed that performance in downstream tasks like POS-tagging and MT is especially hurt. 
\citet{sakaguchi2017robsut} 
studied word scrambling and the \emph{Cmabrigde Uinervtisy (Cambridge University) effect}, where humans are able to understand the meaning of sentences with scrambled words,
performing word recognition (word level spelling correction) with a semi-character RNN system.

Focusing only on character-level NMT models, \citet{belinkov2017synthetic} showed that they exhibit degraded performance when presented with noisy test examples (both artificial and natural occurring noise). 
In line with our findings, they also showed that slightly better performance can be achieved by training on data artificially induced with the same kind of noise as the test set.

\citet{sperbertoward} proposed a noise-introduction system reminiscent of WER, based on insertions, deletions, and substitutions. An NMT system tested on correct transcriptions achieves a BLEU score of $55$ (4 references), but tested on the ASR transcriptions it only achieves a BLEU score of $35.7$. By introducing similar noise in the training data, they were able to make the NMT system slightly more robust. Interestingly, they found that the optimal amount of noise on the training data is smaller than the amount of noise on the test data.

The notion of linguistically plausible corruption is also explored by \citet{li2017robust}, who created adversarial examples with syntactic and semantic noise (reordering and word substitutions respectively). When training with these noisy datasets, they obtained better performance on several text classification tasks. Furthermore, in accordance with our results, their best system is the one that combines different types of noise.

We present a summary of relevant previous work in Table~\ref{tab:previous}. 
\emph{Synthetic} errors refer to noise introduced according an artificially created distribution, and \emph{natural} errors refer to actual errorful text produced by humans. As for \emph{semi-natural}, it refers to either noise introduced according to a distribution learned from data (as in our work), or to errors that are learned from data but introduced according to an artificial distribution (as is part of the work of \citet{belinkov2017synthetic}).

\begin{table}[t]
    \centering
    \begin{tabular}{ccc}
    \toprule
        \multirow{2}{*}{Substitution} & \multicolumn{2}{c}{model BLEU difference} \\ 
        & \clean & \mixall\\
    \midrule
        in$\rightarrow$with & $-6.7$ & $-1.7$ \\
        on$\rightarrow$for & $-6.0$ & $-0.1$ \\
        to$\rightarrow$on & $-2.9$ & $-0.5$ \\
        in$\rightarrow$ {\emp} & $-1.8$ & $-1.9$ \\
        {\emp}$\rightarrow$for & $-1.6$  & $-0.6$ \\
    \midrule
        {\emp}$\rightarrow$\emph{any} & $-1.2$ & $-0.4$ \\
        \emph{any}$\rightarrow${\emp} & $-0.8$ & $-0.7$ \\
    \bottomrule
    \end{tabular}
    \caption{Effect of selected preposition substitutions in test data (\prep) relative to clean test data (\clean), for the \clean and \mixall models. The \mixall model handles most errors more efficiently.}
    \label{tab:prep}
\end{table}

We consider our work to be complementary to the works of \citet{heigold2017robust,belinkov2017synthetic}, and \citet{sperbertoward}. However, there are several important differences:
\begin{table*}[t]
    \centering
    \begin{tabular}{lc|p{4cm}|cc}
    \toprule
        Work & Errors & Noise Types & NMT level & Languages \\
    \midrule
        \cite{heigold2017robust} & synthetic & character swaps, character flips, word scrambling & char, BPE & De$\rightarrow$En\\
    \midrule
        \cite{sperbertoward} & synthetic & ASR errors & word & Es$\rightarrow$En\\
    \midrule
        \multirow{4}{*}{\cite{belinkov2017synthetic}} & synthetic & character swap, middle scramble, full scramble, keyboard typo & \multirow{2}{*}{char, BPE} & \multirow{2}{*}{Fr,De,Cz$\rightarrow$En} \\
        & semi-natural & word substitutions & \\ 
    \midrule
        \multirow{4}{*}{this work} & semi-natural & grammar errors: article, preposition, noun number, verb agreement & \multirow{2}{*}{BPE} & \multirow{2}{*}{En$\rightarrow$Es}\\
        & natural & JFLEG corpus & \\
    \bottomrule
    \end{tabular}
    \caption{Previous work on evaluating the effect of noise in NMT systems. Character swaps refer to neighboring character reordering (e.g. noise$\rightarrow$nosie), while character flips refer to character substitutions (e.g. noise$\rightarrow$noiwe).}
    \label{tab:previous}
\end{table*}
\begin{enumerate}[nolistsep]
    \item \citet{belinkov2017synthetic} and \citet{sperbertoward} train their NMT systems on fairly small datasets: 235K (Fr-En), 210K (De-En), 122K (Cz-En), and 138K sentences (Es-En) respectively. Even though they use systems like Nematus \cite{sennrich2017nematus} or XNMT \cite{neubig2018xnmt} which generally achieve nearly SOTA results, it is unclear whether their results generalize to larger training data. In contrast, we train our system on almost 2M sentences.
    
    \item All three systems introduce somewhat unrealistic amounts of noise in the data. The natural noise of \citet{belinkov2017synthetic} consists of word substitutions based on Wikipedia errors or corrected essays (in the Czech case) but they substitute all possible correct words with their erroneous version, ending up with datasets with more than $40\%$ of the tokens being noisy. For that reason, we refer to it as \emph{semi-natural} noise in Table~\ref{tab:previous}.  Meanwhile, \citet{sperbertoward} test on the outputs of an ASR system that has a WER of $41.3\%$. For comparison, in the \jfleg datasets, we calculated that only about $3.5\%$--$5\%$ of the tokens are noisy -- the average Levenshtein distance of a corrected reference and its noisy source is 13 characters.
    
    \item The word scrambling noise, albeit interesting, could not be claimed to be applicable to realistic scenarios, especially when applied to all words in a sentence. 
    The solution \citet{belinkov2017synthetic} suggested and \citet{sperbertoward} discussed is a character- or spelling-aware model for producing word- or subword-level embeddings. We suspect that such a solution would indeed be appropriate for dealing with typos and other character-level noise, but not for more general grammatical noise. 
    Our method could potentially be combined with GloVe \cite{pennington2014glove} or fastText \cite{bojanowski2017enriching} embeddings that can deal with slight spelling variations, but we leave this for future work.
\end{enumerate}

On the other side, Grammar Error Correction has been extensively studied, with significant incremental advances made recently by treating GEC as an MT task: 
among others, \citet{junczysdowmunt-grundkiewicz:2016:EMNLP2016} used phrased-based MT, \citet{ji-EtAl:2017:Long} used hybrid character-word neural sequence-to-sequence systems, \citet{sakaguchi2017grammatical} used reinforcement learning, and \citet{dowmunt2018etal} combined several techniques with NMT to achieve the current state-of-the-art.
Synthetic errors for training GEC systems have also been studied and applied with mixed success \cite{rozovskaya2010generating,rozovskaya2014illinois,xie2016neural}, while more recently \citet{xie2018noising} used backtranslation techniques for adding synthetic noise useful for GEC.

\section{Conclusion}

In this work, we studied the effect of grammatical errors in NMT.
We not only confirmed previous findings, but also expanded on them, showing that \emph{realistic human-like} noise in the form of specific grammatical errors also leads to degraded performance.
We added synthetic errors on the English \wmt training, dev, and test data (including dev and test sets for all \wmt18 evaluation pairs), and have released them along with the scripts necessary for reproducing them.\footnote{\url{https://bitbucket.com/antonis/nmt-grammar-noise}}
We also produced Spanish translations of the \jfleg{} corpus, so that future NMT systems can be properly evaluated on real noisy data.

\bibliographystyle{acl_natbib}
\bibliography{References}

\appendix

\section{Results with LSTM models}
\begin{table*}[h!]
\newcommand{\match}[1]{\hl{#1}}
    \centering
    \begin{tabular}{c|cccccc|c}
    \toprule
        \multirow{2}{*}{Training Set} &  \multicolumn{7}{c}{En-Es WMT Test Set}\\ 
         &  \clean & \drop & \art & \prep & \nn &\sva & \textsc{average} $\pm$ \textsc{stdev} \\
        \midrule
        \textsc{wmt}-\clean & \textbf{\match{26.62}} & 24.08 & 25.35 & 25.63 & 23.34 & \textbf{26.06} & $25.18 \pm 1.24 $\\
        \textsc{wmt}-\drop & 25.10 & \match{24.21} & 24.24 & 24.00 & 22.26 & 19.58 & $23.23 \pm 2.02 $\\
        \textsc{wmt}-\art  & 25.49 & 23.26 & \match{24.78} & 24.35 & 22.42 & 25.59 & $24.31 \pm 1.26 $\\
        \textsc{wmt}-\prep & 25.49 & 22.99 & 24.39 & \match{25.22} & 22.78 & 25.07 & $24.32 \pm 1.17$\\
        \textsc{wmt}-\nn   & 25.35 & 23.04 & 23.06 & 24.15 & \match{24.73} & 24.61 & $24.16 \pm 0.94$\\ 
        \textsc{wmt}-\sva  & 25.77 & 23.49 & 24.68 & 24.62 & 23.22 & \match{25.41} & $24.53 \pm 1.01$\\ 
    \midrule
        \textsc{wmt}-\dropd & \textbf{26.45} & \textbf{\match{25.37}} & 25.59 & 25.59 & 23.64 & 25.92 & $25.43 \pm 0.95$\\
        \textsc{wmt}-\artd  & \textbf{26.64} & 24.60 & \textbf{\match{26.35}} & 26.08 & 23.69 & 26.48 & \textbf{25.64 $\pm$ 1.21}\\
        \textsc{wmt}-\prepd & \textbf{26.60} & 24.31 & 25.12 & \textbf{\match{26.30}} & 23.27 & \textbf{26.14} & $25.29 \pm 1.31$\\
        \textsc{wmt}-\nnd & 26.23 & 23.86 & 24.75 & 25.52 & \textbf{\match{25.20}} & 25.66 & $25.20 \pm 0.82$\\ 
        \textsc{wmt}-\svad  & \textbf{26.62} & 24.22 & 25.49 & 25.86 & 23.79 & \textbf{\match{26.24}} & $25.37 \pm 1.13$\\ 
    \midrule
        \textsc{wmt}-\mixd 
        & \textbf{26.60} & 24.90 & 25.52 & 25.80 & 24.68 & \textbf{26.03} & \textbf{25.59 $\pm$ 0.72} \\
    \bottomrule
    \end{tabular}
    \caption{BLEU scores on the WMT test set without (\textsc{clean}) and with synthetic grammar errors using an LSTM encoder-decoder model.}
    \label{tab:results-lstm}
\end{table*}

\iftrue
\begin{table*}[h!]
    \centering
    \jfleges Dev \\[1ex]
    \begin{tabular}{c|ccccc|cc}
    \toprule
        \multirow{2}{*}{Training} &  \multicolumn{5}{c|}{Manual correction} & No & Auto \\ 
         & \rf0 & \rf1 & \rf2 & \rf3 & avg. & corr. & corr. \\
    \midrule
        \clean & \textbf{28.3} & 27.3 & 28.4 & 28.2 & 28.0 & 27.1 & 27.7 \\
        \mixd  & 28.2 & \textbf{27.5} & \textbf{28.8} & \textbf{29.1} & \textbf{28.4} & \textbf{27.4} & \textbf{28.2}\\
    \bottomrule
    \end{tabular}
    
    \vspace*{2ex}
    \jfleges Test \\[1ex]
    \begin{tabular}{c|ccccc|cc}
    \toprule
        \multirow{2}{*}{Training} &  \multicolumn{5}{c|}{Manual correction} & No & Auto \\ 
         & \rf0 & \rf1 & \rf2 & \rf3 & avg. & corr. & corr. \\
    \midrule
    \clean & \textbf{24.9} & \textbf{25.1} & \textbf{25.6} & \textbf{25.1} & \textbf{25.2} & 22.8 & 23.5 \\
    \mixd & 24.8 & 25.0 & 25.3 & 25.0 & 25.0 & \textbf{23.1} & \textbf{24.3} \\
    \bottomrule
    \end{tabular}
    \caption{BLEU scores on the \jfleges dev and test datasets with the LSTM encoder-decoder model.}
    \label{tab:jfleg-results-test-lstm}
\end{table*}

\fi

\end{document}